\documentclass[11pt]{article}
\usepackage{graphicx}
\usepackage{url}
\usepackage{hyperref}
\usepackage[numbers, sort&compress]{natbib}
\usepackage{fancyhdr}
\usepackage{colm2024_conference}
\usepackage[utf8]{inputenc}
\usepackage[T1]{fontenc}
\usepackage{booktabs}
\usepackage{amssymb}
\usepackage{amsmath}
\usepackage{colortbl}
\usepackage{tabularx}
\usepackage{array}
\usepackage{geometry}
\usepackage{xspace}
\usepackage{enumitem}
\usepackage{float}

\usepackage{multirow}
\usepackage{makecell}
\usepackage[table]{xcolor}
\usepackage{threeparttable}
\usepackage{wrapfig}
\newcolumntype{L}{>{\raggedright\arraybackslash}X}
\newcolumntype{M}{>{\raggedright\arraybackslash}p{3.5cm}}

\usepackage{hyperref}
\definecolor{darkblue}{rgb}{0, 0, 0.5}
\definecolor{skyblue}{rgb}{0.289, 0.484, 0.72}
\hypersetup{colorlinks=true, citecolor=skyblue, linkcolor=red, urlcolor=darkblue, bookmarksopen=false, bookmarksnumbered=true}

\definecolor{groupgray}{RGB}{241,241,241}
\definecolor{oursred}{RGB}{255,232,232}

\definecolor{mygreen}{RGB}{0,160,0}   
\definecolor{myred}{RGB}{200,0,0}     
\newcommand{\yes}{\textcolor{mygreen}{$\checkmark$}}
\newcommand{\no}{\textcolor{myred}{$\times$}} 
\usepackage{graphicx}
\newcolumntype{S}[1]{>{\centering\arraybackslash}m{#1}} 
\newcolumntype{L}{>{\raggedright\arraybackslash}X}
\newcolumntype{C}{>{\centering\arraybackslash}X}

\newcolumntype{P}[1]{%
  >{\raggedright\arraybackslash}p{#1}}
\newcolumntype{Y}{%
  >{\raggedright\arraybackslash}X}
  
\newcommand{\ours}{CPI-Bench\xspace}

\title{\ours: A Comprehensive, Practical and Intelligent Benchmark for Real-World Image Editing}
\author{
Qinye Zhou, Jun Zheng, Yongchao Du, Yuan Wang, Zhengrui Chen, Zuan Gao, Taihang Hu, Chao Lin, Yefeng Shen,  Xingjian Wang, Zhao Wang, Zhengtao Wu, Xiaoli Xu, Zhengze Xu, Hao Yan, Denghui Yang, Yuhang Yu, Huayu Zhang, Mingzhou Zhang,  Mengting Chen$^{\dagger}$ \\ \vspace{2mm}
\textbf{Alibaba Group} \\ \vspace{2mm}
{\small  $^{\dagger}$Corresponding Author}
}

\begin{document}
\maketitle
\begin{abstract}
With the rapid advancement of image editing models and their widespread application across various domains, there is an increasingly urgent need to deploy these model capabilities directly into real-world scenarios. However, existing benchmarks remain confined to simple single-image tasks, suffering from limited coverage dimensions and an inability to effectively differentiate performance among diverse models. Consequently, they fail to reliably evaluate model performance in complex multi-image editing, highly demanding reasoning instructions, and practical deployment settings. To address these limitations, we propose \textbf{CPI-Bench}, a \textbf{C}omprehensive,  \textbf{P}ractical and \textbf{I}ntelligent benchmark for real-world image editing. CPI-Bench comprises three core subsets: CPI-General-Bench, which comprehensively covers diverse editing tasks and introduces multi-image editing evaluation; CPI-Practical-Bench, which focuses on high-frequency real-user application scenarios; and CPI-Intelligent-Bench, which is dedicated to evaluating capabilities in highly demanding reasoning-based editing. Evaluation results of mainstream image editing models based on CPI-Bench demonstrate that CPI-Bench enhances performance differentiation among models. It provides a comprehensive and reliable quantification of gaps in general editing capabilities, practical deployment efficacy, and advanced reasoning-based editing, offering invaluable guidance for the future optimization of image editing models. Crucially, our ranking analysis reveals that CPI-Bench achieves the highest alignment with the Arena Image Edit Leaderboard, indicating stronger consistency with public human preference rankings, serving as an effective proxy for public human evaluations.

\end{abstract}

\begin{figure}[H]
    \centering
     \includegraphics[width=1.0\textwidth]{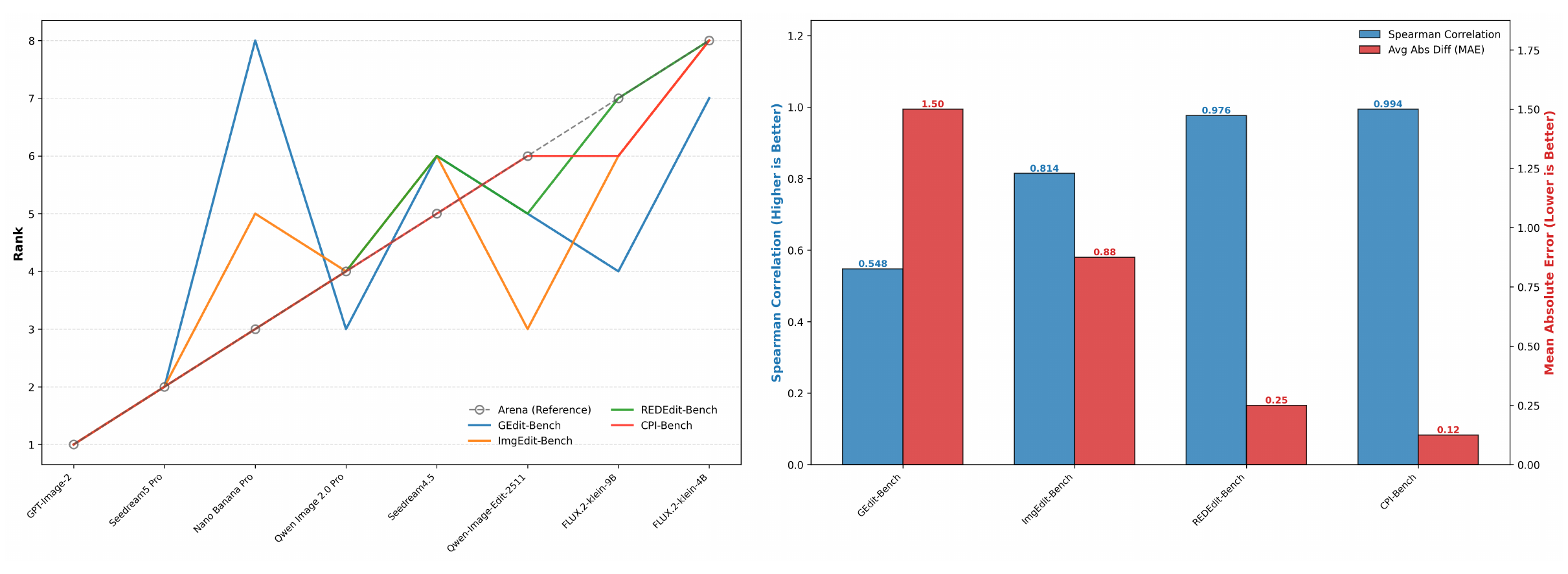}
  \caption{\textbf{Left:} Model ranking trends across benchmarks compared against the Arena Image Edit Leaderboard~\cite{lmarena2024leaderboard}.  Our proposed CPI-Bench demonstrates the closest alignment with Arena. \textbf{Right:} Spearman correlation coefficients and Mean Absolute Error (MAE) with Arena rank. CPI-Bench outperforms other benchmarks, achieving the highest correlation and the lowest error margin.}
  \label{fig:rank_compare}
\end{figure}


\begin{figure}
    \centering
    \includegraphics[width=0.95\linewidth]{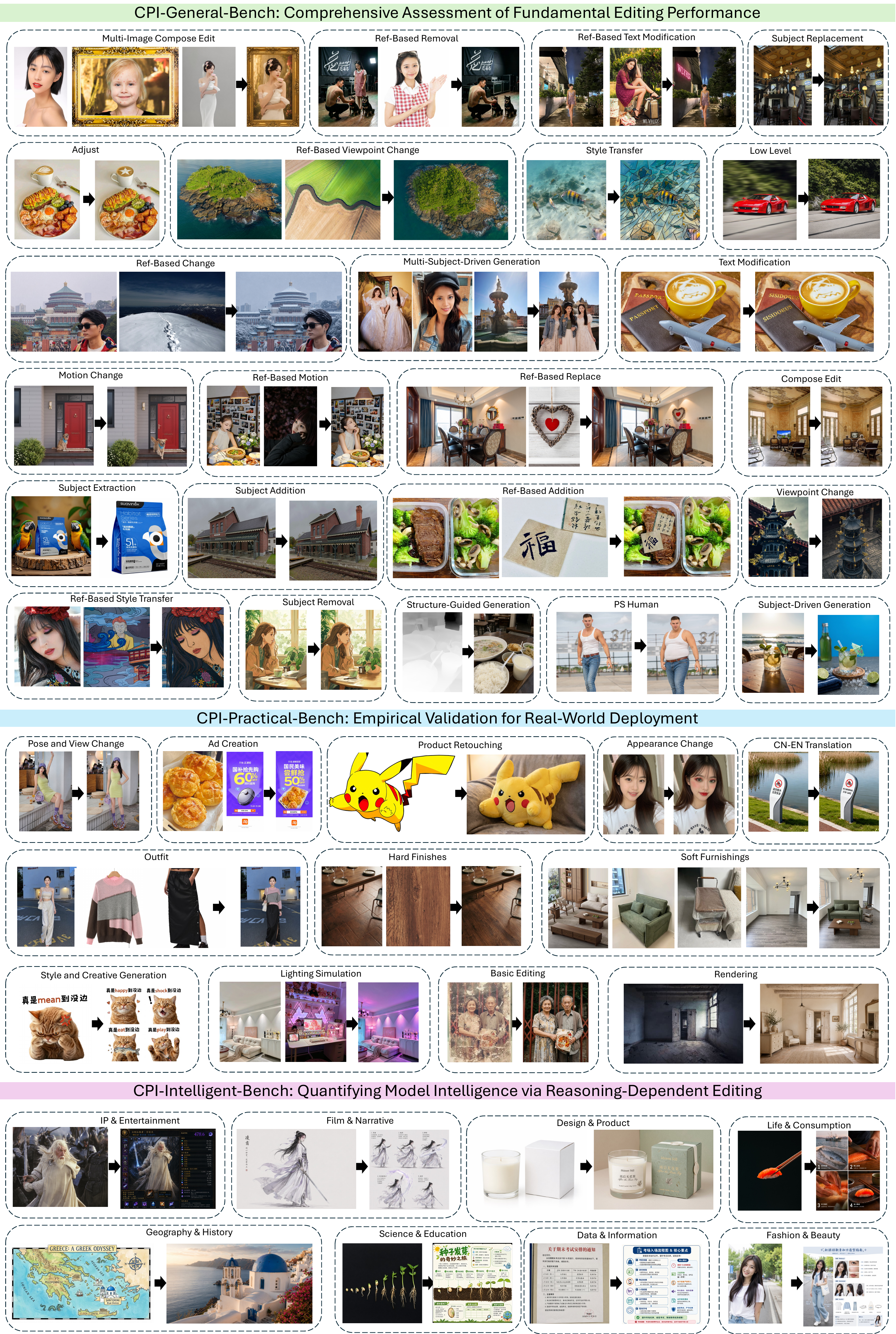}
  \caption{The overview of CPI-Bench. CPI-Bench is decomposed into three distinct dimensions: CPI-General-Bench for comprehensive editing performance, CPI-Practical-Bench for real-world deployment efficacy, and CPI-Intelligent-Bench for reasoning-based editing tasks.}
  \label{fig:overview}
\end{figure}
\section{Introduction}

In recent years, image generation technologies have witnessed remarkable progress\citep{ho2020denoising, song2020denoising, rombach2022high, peebles2023scalable, esser2024scaling}. Top-tier closed-source models (e.g., GPT-Image2\citep{gptimage2_model_card}, Nano Banana Pro\citep{google2025nanobanana}) now possess the capability to perform complex image editing guided by complex interleaved image-text instructions.  The application scenarios of image editing models are rapidly shifting from simple trials to real-world deployment, with users increasingly eager to integrate these models directly into practical production environments. However, relative to the rapid evolution of model capabilities, existing evaluation systems lag significantly behind. Current mainstream benchmarks are often confined to simple single-image editing tasks, failing to reliably reflect performance disparities among different models in complex multi-image editing, real-world deployment scenarios, and highly demanding reasoning-based editing. This lack of evaluative dimensions not only hinders users from making scientifically grounded model selections based on actual needs but also restricts researchers from precisely identifying model defects and defining clear optimization directions.

Although existing image editing benchmarks (e.g., GEdit-Bench\cite{liu2025step1x-edit}, ImgEdit-Bench\cite{ye2025imgedit}, REDEdit-Bench\cite{firered2026rededit} ) have been widely adopted, they suffer from significant limitations:
First, they have incomplete task coverage and low difficulty. Existing benchmarks primarily focus on basic single-image tasks, lacking consideration for high-difficulty editing operations. For instance, GEdit-Bench and ImgEdit-Bench omit tasks involving viewpoint change, while REDEdit-Bench lacks critical capabilities such as single-subject-driven editing and subject Re-orientation. More critically, these benchmarks completely overlook multi-image editing tasks. In modern image generation and editing workflows, maintaining consistency across multiple images and enabling cross-image interactive editing are key indicators of a model's core competence.
Second, a lack of real-world scenario orientation. Existing evaluations categorize tasks based solely on abstract capability dimensions, detached from authentic application contexts. This design renders evaluation results ineffective at reflecting model performance in practical settings such as daily life and consumer applications, failing to quantify the true deployment value of models and creating a disconnect between academic metrics and practical utility.
Third, existing general image editing benchmarks largely lack data on highly demanding reasoning-based editing. This omission results in a critical gap in evaluating such capabilities, rendering it impossible to assess the true intelligence level of these models.

\begin{figure}[t]
    \centering
    \includegraphics[width=0.95\linewidth]{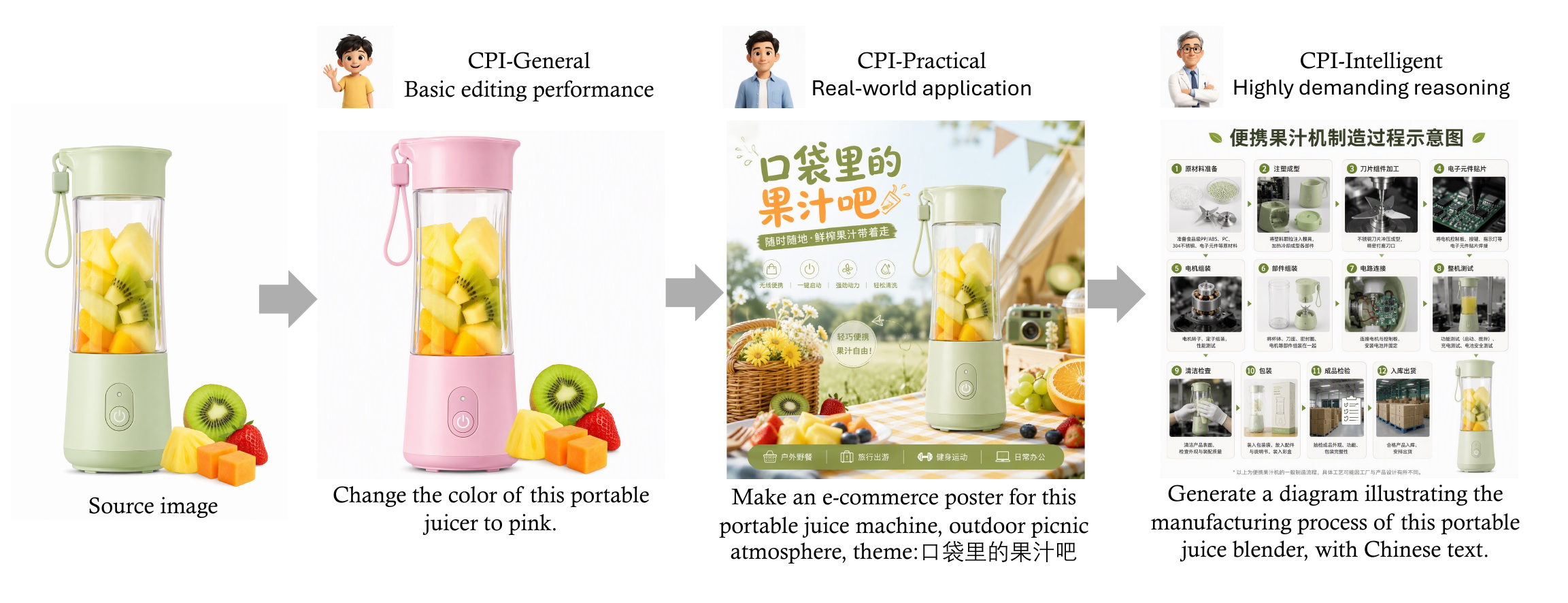}
  \caption{Hierarchical evaluation of model capabilities across CPI-Bench:
(a) Basic editing performance (CPI-General); (b) Real-world application robustness (CPI-Practical); and (c) Highly demanding reasoning tasks (CPI-Intelligent). The difficulty escalates progressively from general, practical to intelligent levels.}
  \label{fig:three_level}
\end{figure}

To address these challenges, we propose CPI-Bench, a comprehensive, practical and intelligent benchmark for image editing. As shown in Figure~\ref{fig:overview}, CPI-Bench consists of three complementary subsets:
\textbf{(1). CPI-General-Bench:} It focuses on a comprehensive assessment of fundamental editing capabilities. This subset covers 24 diverse editing tasks comprising 2,039 evaluation samples. Notably, it includes 14 single-image tasks and 10 multi-image tasks, effectively filling the void in current evaluations of multi-image editing capabilities.
\textbf{(2). CPI-Practical-Bench:} It is a cross-domain benchmark to systematically evaluate image editing performance across diverse real-world consumer scenarios. We constructed a dataset covering 51 common application types with 558 samples, spanning four core domains: portrait enhancement, e-commerce \& advertising creativity, residential \& interior design, and content creation (e.g., ID photo generation, IP product rendering, virtual furniture placement and multi-panel story generation). This achieves a paradigm shift from academic capability assessment to practical deployment evaluation.
\textbf{(3). CPI-Intelligent-Bench:} It focuses on evaluating the capabilities of image editing models in highly demanding reasoning-based editing. CPI-Intelligent-Bench is constructed by optimizing and curating the dataset from our prior work, ExpertVerse \cite{wang2026expertversegeneralpurposebenchmarkexpertlevel}, comprising 1,181 reasoning instances spanning 8 expert domains and 67 sub-disciplines.  The difficulty of these three subsets escalates progressively from general, practical to intelligent levels  as shown in Figure~\ref{fig:three_level}.

Furthermore, we designed an automated evaluation framework based on Vision-Language Models (VLMs). For each task category, we customized specific scoring prompts to quantitatively assess results across three dimensions: Instruction Adherence, Visual Naturalness, and Physical \& Detail Consistency. We conducted extensive evaluations of mainstream open-source and closed-source image editing models on CPI-Bench. Experimental results demonstrate that, compared to traditional benchmarks, CPI-Bench significantly amplifies performance discrepancies between models. It more acutely captures model shortcomings in multi-image, real-world and highly demanding reasoning scenarios, thereby providing clear guidance for future model optimization. Furthermore, to  validate the alignment between our benchmarks and human consensus, we conducted a comparative analysis of model rankings across all evaluated benchmarks against the Arena Image Edit Leaderboard. As illustrated in Figure~\ref{fig:rank_compare}, the ranking distribution derived from CPI-Bench exhibits the tightest correlation with the Arena, significantly outperforming other existing benchmarks. This superior alignment provides compelling evidence that CPI-Bench's evaluation outcomes show strong alignment with the perceptual preferences reflected in Arena rankings.

In summary, the main contributions of this paper are as follows:

\begin{itemize}[leftmargin=*]
\item \textbf{[Construction of a Comprehensive Task Coverage System, Filling the Multi-Image Evaluation Gap.]} We propose CPI-General-Bench, whose general subset encompasses 24 broad and diverse editing tasks. Crucially, by introducing 10 multi-image editing tasks, we address the long-standing neglect of multi-image interaction capabilities in existing benchmarks, providing a standardized testbed for evaluating complex reasoning and consistency maintenance.
\item \textbf{[Introducing a Deployment-Oriented Benchmark for Real-World Consumer Scenarios.]}  We introduce the CPI-Practical-Bench, a cross-domain dataset focused on practical application outcomes. By curating 51 typical tasks across four major domains, we elevate the evaluation perspective from singular academic metrics to authentic user experiences, achieving a comprehensive quantification of model deployment feasibility.
\item \textbf{[Full-spectrum Evaluation for Image Editing Model]}  CPI-Bench offers a holistic performance assessment of image editing models across three critical dimensions: general editing capability, practical deployment efficacy, and model intelligence. 

\item \textbf{[Revelation of True Performance Gaps in Complex Scenarios.] } Experiments demonstrate that CPI-Bench provides stronger model discrimination than existing benchmarks limited to simple single-image tasks. On CPI-Bench, the performance variance among different models is significantly enhanced, clearly characterizing the capability hierarchy of models in handling highly demanding reasoning instructions, multi-image correlations, and real-world applications, thus offering insights for future model development.

\item \textbf{[Alignment with Human Preferences via Arena Correlation.] } Our ranking analysis reveals that CPI-Bench achieves the highest statistical alignment with the Arena Image Edit Leaderboard among all evaluated benchmarks. This strong correlation confirms that CPI-Bench does not merely measure technical metrics but shows strong alignment with the  preferences of human evaluators. Consequently, it serves as an effective, reliable proxy for public preference trends, ensuring that model optimization is better aligned with public human evaluations.
\end{itemize}

\section{Related work}
\subsection{Image Editing Benchmark}
Although widely utilized, existing general benchmarks\citep{liu2025step1x-edit,firered2026rededit,ye2025imgedit} remain constrained by narrow task coverage and a predominant focus on single-image settings. Their taxonomies rely heavily on abstract capability dimensions, often neglecting the nuances of real-world deployment. 
Most significantly, these benchmarks completely ignore the crucial domain of multi-image editing.

Furthermore, existing general image editing benchmarks lack reasoning-based evaluation data, rendering them incapable of assessing the intelligence of editing models. Conversely, current reasoning-specific benchmarks are overly singular in their focus on cognitive capabilities and exhibit distinct limitations in scope: RISEBench~\cite{zhao2026envisioning} primarily evaluates temporal, causal, spatial, and logical dimensions; KRISBench~\cite{wu2026kris} introduces a knowledge-grounded taxonomy covering factual, conceptual, and procedural types but suffers from insufficient knowledge depth; UniREditBench~\cite{han2025unireditbench} expands the scope through game-world scenarios and multi-object interactions; while WiseEdit~\cite{pan2026wiseedit} assesses cognition- and creativity-informed editing via a three-stage pipeline of awareness, interpretation, and imagination. To bridge these gaps, we propose CPI-Bench, a comprehensive framework capable of holistically evaluating image editing models across three critical dimensions: general editing capabilities, practical deployment in real-world scenarios, and intelligence of reasoning-based editing.

\subsection{Image Editing  Evaluation Methods}

Traditional evaluation approaches\citep{kawar2023imagic, ma2024i2ebench, li2024seed, hui2024hq,korhonen2012peak,wang2023imagen} primarily depended on generic similarity metrics (e.g., CLIP Score~\cite{radford2021learning}, PSNR~\cite{korhonen2012peak}, SSIM~\cite{wang2004image}), which frequently demonstrate limited correlation with human perceptual judgments. In response, recent advancements (such as GEdit-Bench\cite{liu2025step1x-edit},ImgEdit-Bench\cite{ye2025imgedit}) have pivoted toward utilizing Vision-Language Models (VLMs) as evaluators, thereby substantially enhancing consistency with human preferences. Building upon this foundation, we propose a VLM-based automated evaluation framework. Our method engineers task-specific scoring prompts that adapt to the unique nuances of each editing category, facilitating a more fine-grained and precise evaluation process.
\section{CPI-Bench}

\begin{figure}[t]
\centering
  \includegraphics[width=1.0\linewidth]{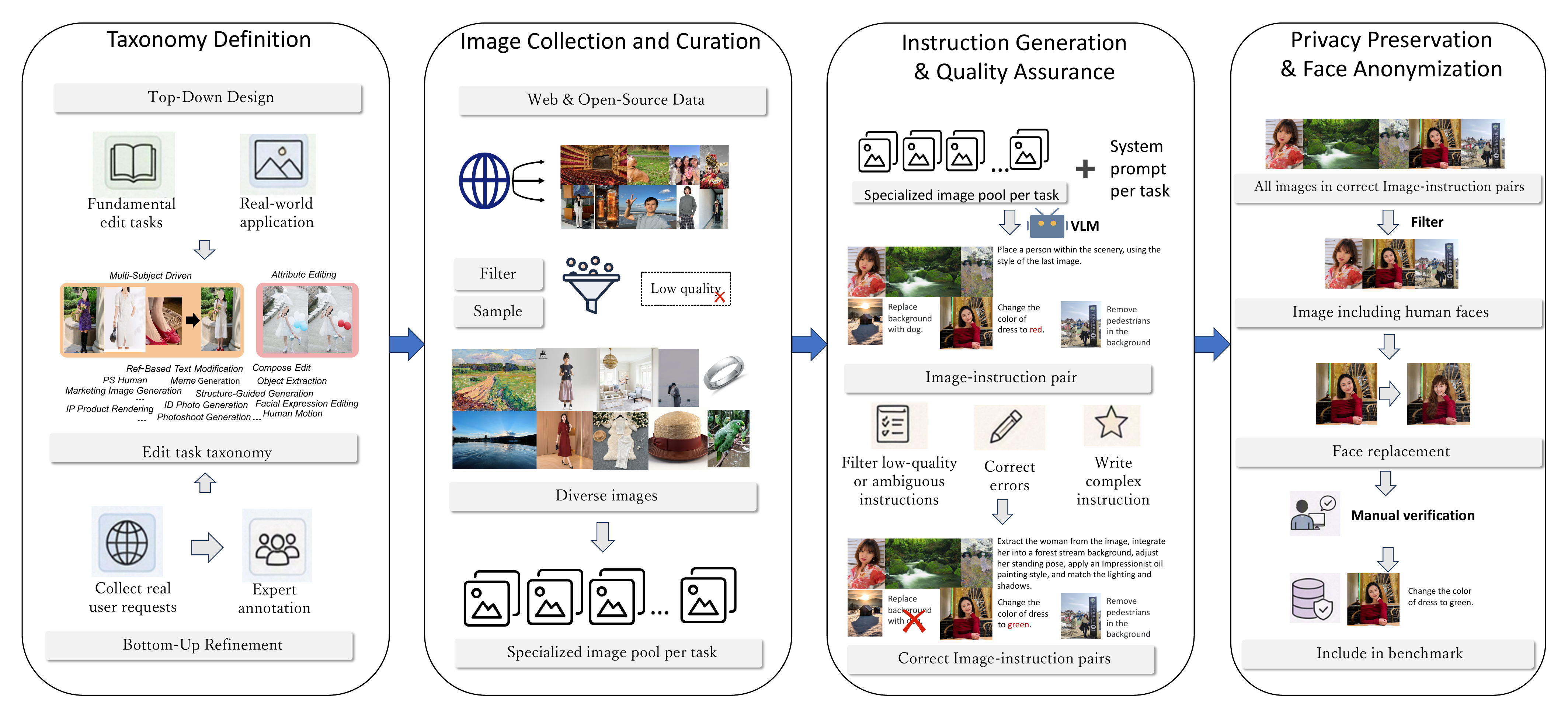}
  \caption{The construction pipeline of CPI-Bench. The process comprises four sequential stages: (1) Taxonomy Definition, (2) Image Collection and Curation, (3) Instruction Generation and Quality Assurance, and (4) Privacy Preservation and Face Anonymization.}
  \label{fig:construction_pipeline}
\end{figure}

\begin{figure}[t]
\centering
  \includegraphics[width=1.0\textwidth]{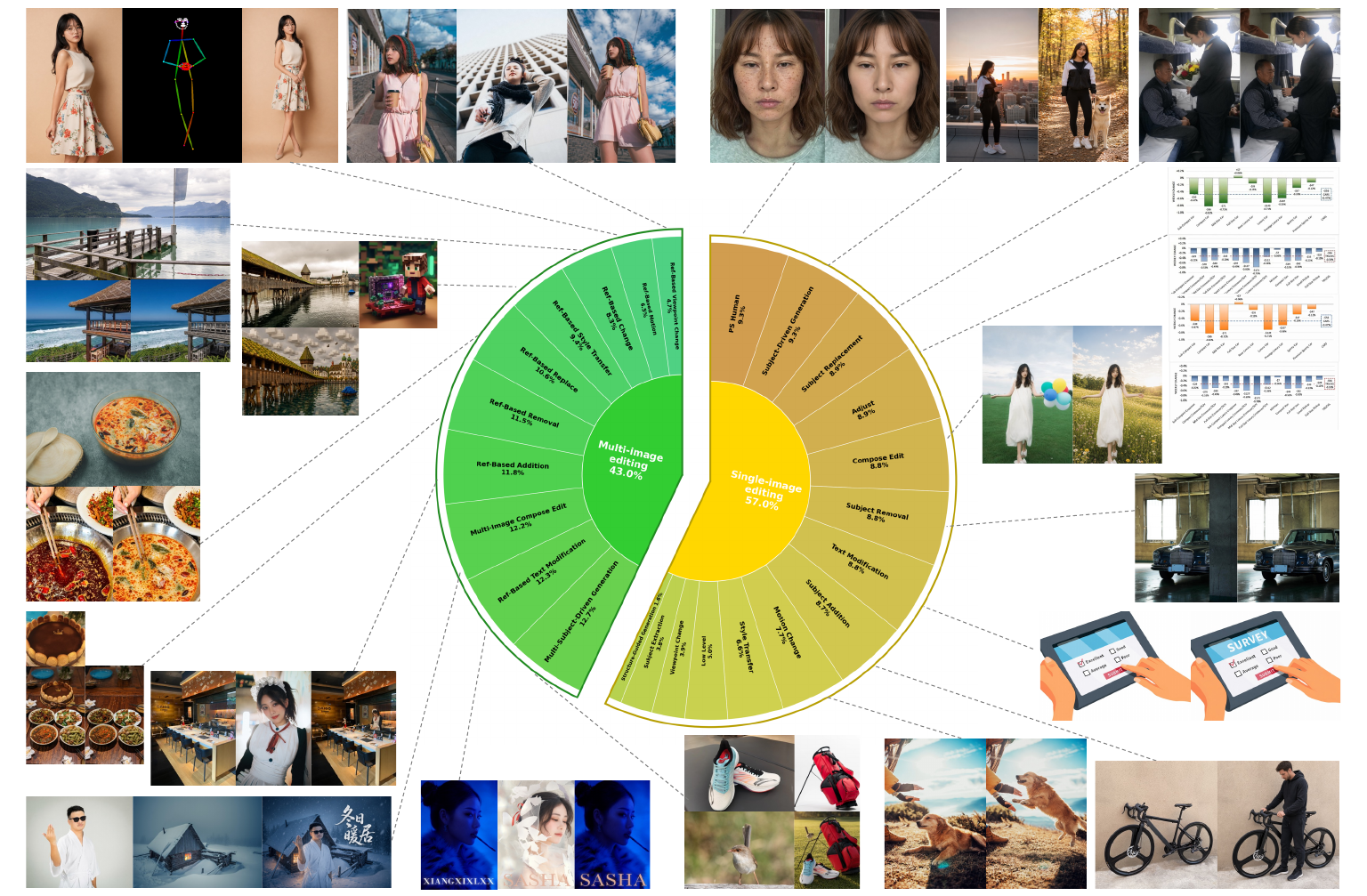}
  \caption{CPI-General-Bench distribution. CPI-General-Bench comprises 14 single-image editing tasks and 10 multi-image editing tasks.}
  \label{fig:general_bench_distribution}
\end{figure}

\begin{figure}[t]
\centering
  \includegraphics[width=1.0\textwidth]{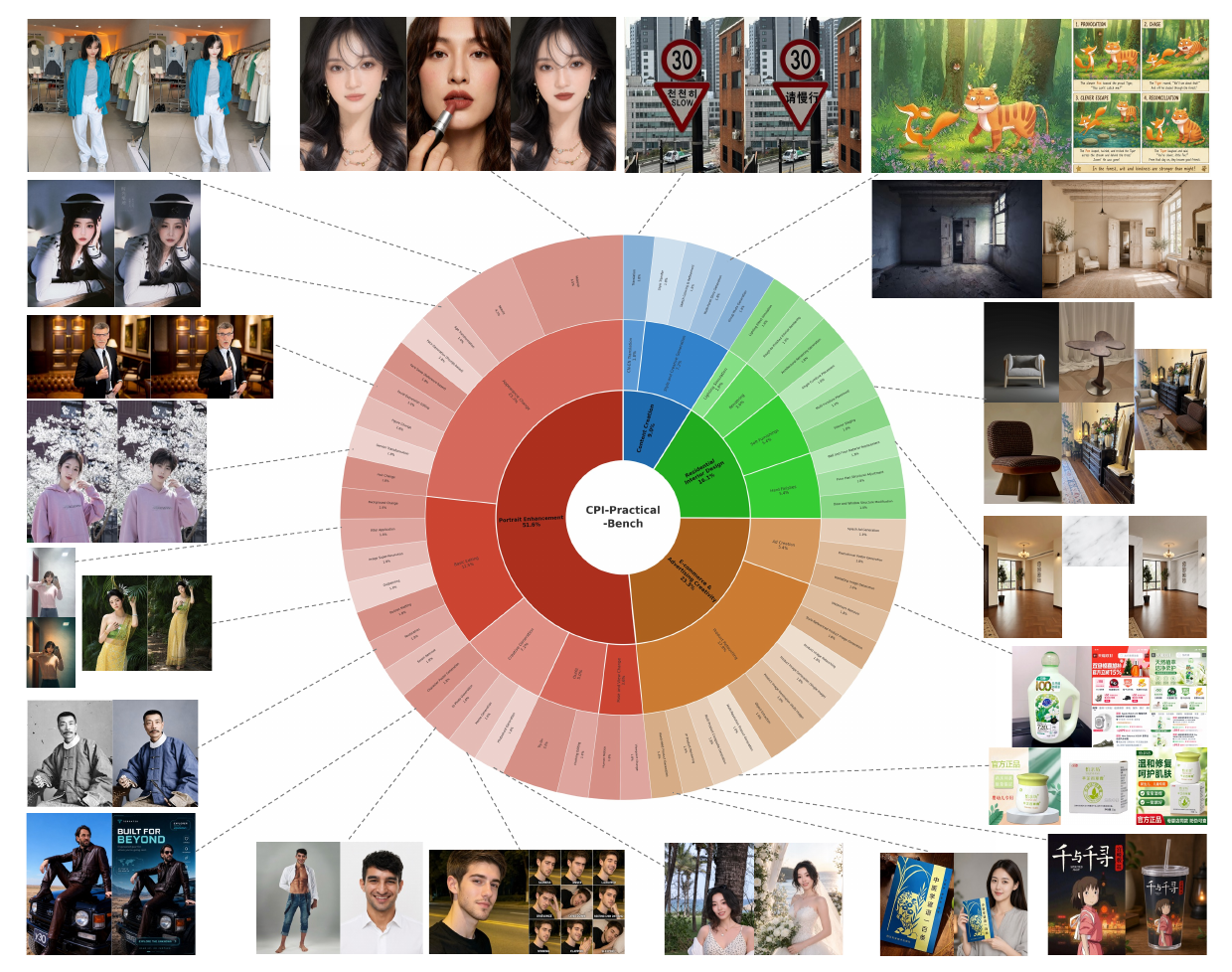}
  \caption{CPI-Practical-Bench distribution. CPI-Practical-Bench comprises 51 common application scenarios, covering 4 top-level categories and 13 sub-categories.}
  \label{fig:practical-bench_distribution}
\end{figure}

In this section, we first introduce the construction pipeline of CPI-Bench in Section~\ref{section:3.1}. It is worth noting that CPI-Intelligent-Bench builds upon our prior work, ExpertVerse \cite{wang2026expertversegeneralpurposebenchmarkexpertlevel}. As it adopts the same data construction methodology and the only modification is that we further refined the dataset by filtering and optimizing samples that failed to meet specific reasoning criteria, we omit a repeated description of its construction pipeline here and instead focus on CPI-General-Bench and CPI-Practical-Bench.  Subsequently, Section~\ref{section:3.2} presents a detailed analysis of the data distribution within CPI-Bench, followed by Section~\ref{section:3.3}, which defines the evaluation metrics.

\subsection{CPI-Bench Construction Pipeline}\label{section:3.1}

To ensure the comprehensiveness, diversity, and ethical compliance of CPI-General-Bench and CPI-Practical-Bench, we established a rigorous four-stage construction pipeline as shown in Figure~\ref{fig:construction_pipeline}:

 \textbf{1. Taxonomy Definition: A Hybrid Top-Down and Bottom-Up Approach.}  
We adopted a dual strategy combining top-down theoretical deduction with bottom-up empirical induction to construct our task taxonomy.
\begin{itemize}[leftmargin=*]
\item Top-Down Design:  Initially, we systematically planned the initial set of editing categories based on two core dimensions: fundamental editing capabilities and real-world application scenarios. This ensured comprehensive coverage of standard editing functions.
\item Bottom-Up Refinement:  Recognizing that purely theoretical frameworks might overlook emerging or long-tail user needs, we introduced a data-driven supplementation mechanism. We extensively collected authentic image editing requests from diverse internet sources and invited domain experts to perform cluster analysis and classification labeling on these demands. Through this iterative process, we successfully identified and filled gaps in the initial framework, ultimately establishing a robust task taxonomy that aligns with both academic standards and actual user intent.
\end{itemize}

 \textbf{2. Image Collection and Curation.}   Based on the defined taxonomy, we meticulously curated a high-quality image repository comprising over 100,000 images sourced from public web datasets and legally licensed commercial channels. The collection spans a wide spectrum of domains, including portraits, animals, plants, natural landscapes, and stylized artistic works.
To guarantee data diversity, we constructed dedicated image pools for each specific editing task. During this process, we strictly controlled the distribution of variables such as background complexity and subject characteristics. This ensured a high degree of heterogeneity within each pool, thereby preventing models from overfitting to specific visual patterns and preserving the generalizability of the evaluation.

 \textbf{3. Instruction Generation and Quality Assurance.}  We employed a Human-in-the-Loop workflow to generate high-fidelity instruction-image pairs:

\begin{itemize}[leftmargin=*]
\item (a) Automated Generation: For each editing category, we designed customized system prompts to drive Vision-Language Models (VLMs). By integrating these prompts with the corresponding image pools, we batch-generated a diverse set of candidate editing instructions.
\item (b) Manual Filtering and Augmentation:   The generated pairs underwent multiple rounds of human review. Annotators filtered out low-quality or semantically ambiguous instructions, corrected logical errors, and manually authored complex, high-difficulty instructions for challenging scenarios to compensate for the limitations of automated generation.

\item (c) Expert Validation:   Finally, all data pairs were subjected to rigorous multi-round inspections by professional technical reviewers. This step ensured the logical consistency, executability, and diversity of the instructions, establishing a high-quality dataset.
\end{itemize}

 \textbf{4. Privacy Preservation and Face Anonymization.}  
To address critical privacy and ethical concerns, we implemented a strict anonymization protocol for all images containing human faces within the benchmark. Utilizing advanced face swapping technology, we replaced original facial features with synthetic, non-identifiable identities while rigorously preserving consistency in expression, pose, and lighting. All anonymized images underwent a secondary manual verification process to confirm accuracy before being officially included in the benchmark. This approach effectively safeguards user privacy while maintaining the utility and realism of the data.
\begin{table*}[t]
    \centering
    \caption{
    Comparison of image editing models across three public benchmarks and our proposed CPI-Bench suite. For multilingual benchmarks, we report results on the English split.
    }
    \label{tab:editing_capability}

    \resizebox{0.99\linewidth}{!}{
    \begin{threeparttable}
    \small
    \setlength{\tabcolsep}{3pt}
    \renewcommand{\arraystretch}{1.2}

    \begin{tabular}{l| c c c c c c c c}
        \toprule
        \textbf{Model}
        & \textbf{Params.}
        & \textbf{GEdit}
        & \textbf{ImgEdit}
        & \textbf{REDEdit}
        & \textbf{CPI-General}
        & \textbf{CPI-Practical}
         & \textbf{CPI-Intelligent}
        & \textbf{CPI-Overall$\uparrow$}$^{\dagger}$
        \\
        \midrule
        \rowcolor{groupgray}[0pt][0pt]
        \multicolumn{9}{@{}l@{}}{
            \hspace{0.5em}\textit{\textbf{Proprietary / Closed-source Models}}
        } \\

        GPT-Image-2\cite{gptimage2_model_card}
        & --
        & 8.69
        & 4.74
        & 4.65
        & 4.64
        & 4.69
        & 4.77
        & 4.70
        \\

        Nano Banana Pro\cite{google2025nanobanana}
        & --
        & 7.73
        & 4.37
        & 4.42
        & 4.47
        & 4.58
        & 4.68 
        & 4.58
        \\

        Seedream5 Pro\cite{seedream2025seedream}
        & --
        & 8.63
        & 4.57
        & 4.62
        & 4.63
        & 4.72
        & 4.70 
        & 4.68
        \\
        
        Seedream4.5\cite{seedream2025seedream}
        & --
        & 7.82
        & 4.32
        & 4.20
        & 4.32
        & 4.33
        & 3.79 
        & 4.15
        \\

        Qwen Image 2.0 Pro\cite{zhao2026qwen}
        & --
        & 8.52
        & 4.45
        & 4.38
        & 4.38
        & 4.39
        & 3.79
        & 4.19 \\
        \midrule
        \rowcolor{groupgray}[0pt][0pt]
        \multicolumn{9}{@{}l@{}}{
            \hspace{0.5em}\textit{\textbf{Open-source Models}}
        } \\

        Qwen-Image-Edit-2511~\cite{wu2025qwen}
        & 20B
        & 7.87
        & 4.51
        & 4.23
        & 3.73
        & 3.85
        & 2.54 
        & 3.37
        \\

        
        
        FLUX.2-klein-9B~\cite{blackforest2025flux2klein}
        & 9B
        & 8.12
        & 4.32
        & 4.07
        & 3.78
        & 3.86
        & 2.48
        & 3.37
        \\
        
        FLUX.2-klein-4B~\cite{blackforest2025flux2klein}
        & 4B
        & 7.79
        & 4.12
        & 3.94
        & 3.65
        & 3.63
        & 2.33 
        & 3.20
        \\
        
        FireRed-Image-Edit~\cite{firered2026rededit}
        & 20B
        & 7.94
        & 4.56
        & 4.26
        & 3.76
        & 3.89
        & 2.65
        & 3.43
        \\

        
        
        JoyAI-Image-Edit-Plus~\cite{song2026joyai}
        & 16B
        & 7.54
        & 4.02
        & 3.87
        & 3.32
        & 3.50
        & 2.07 
        & 2.96
        \\
        

        
        

        


\midrule
\textbf{Variance}
        & -
        & 0.0634
        & 0.0393
        & 0.0585
        & 0.1428
        & 0.1265
        & 0.7282 
        & 0.2876
        \\
        \bottomrule
    \end{tabular}

    \begin{tablenotes}[flushleft]
        \footnotesize
        \item[*] The parameter counts of proprietary models are not publicly disclosed. Proprietary models are evaluated through their official APIs.
    
        \item[$\dagger$] The CPI-Overall score is the unweighted arithmetic mean of the score of CPI-General, CPI-Practical and CPI-Intelligent benchmark listed in the table. Since GEdit uses a 10-point scale while the other benchmarks use a 5-point scale, its score is divided by two before calculating variance. 
    \end{tablenotes}
    \end{threeparttable}}
\end{table*}

\subsection{CPI-Bench Statistics}\label{section:3.2}

As shown in Table~\ref{tab:benchmark_comparison}, compared to prior benchmarks, CPI-Bench offers a significantly broader coverage of editing tasks across capability dimensions. The CPI-General-Bench subset comprises 24 sub-tasks, uniquely incorporating multi-image editing scenarios, bilingual prompts, and task-specific evaluation prompts. Furthermore, the CPI-Practical-Bench, which focuses on assessing performance in real-world application scenarios, encompasses 51 distinct editing applications and CPI-Intelligent-Bench has 67 sub-tasks for evaluating highly demanding reasoning-based editing.

\begin{wrapfigure}{r}{0.25\linewidth}
    \centering
   \vspace{-0.5cm}
    \includegraphics[width=\linewidth]{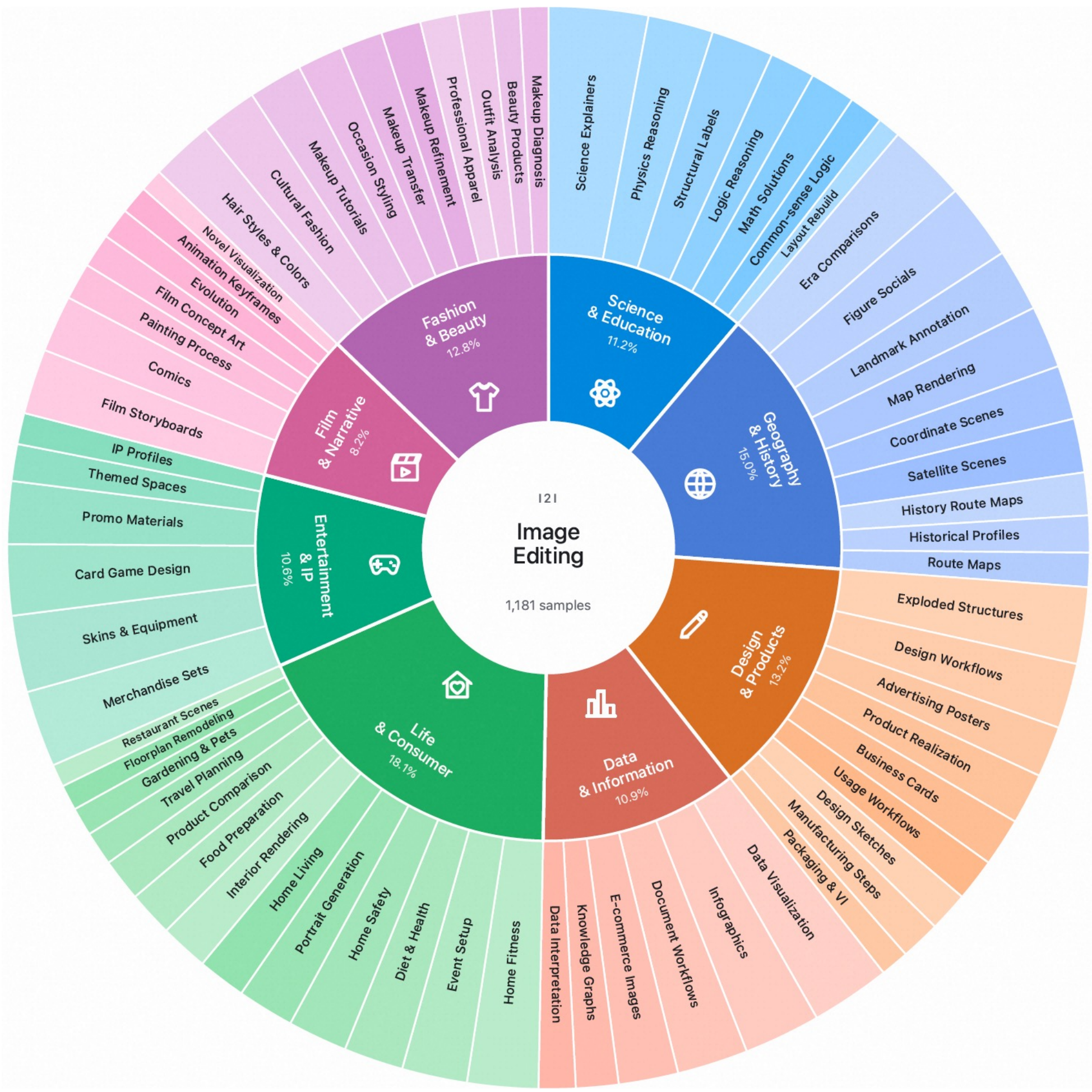}
    \caption{Task distribution in CPI-Intelligent-Bench.}
    \label{fig:Intelligent_bench}
\end{wrapfigure}

As shown in Figure~\ref{fig:general_bench_distribution}, CPI-General-Bench  includes 14 single-image editing tasks and 10 multi-image editing tasks, with a balanced distribution across all sub-tasks. Beyond fundamental tasks common in previous benchmarks (e.g., addition, removal), CPI-General-Bench includes tasks such as subject-driven generation and viewpoint transformation, ensuring comprehensive task coverage. As shown in Figure~\ref{fig:practical-bench_distribution}, CPI-Practical-Bench employs a three-level hierarchical taxonomy consisting of 4 primary categories, 13 secondary types, and 51 tertiary tasks. These span four core domains: Portrait Enhancement, Content Creation, Residential Interior Design, and E-commerce \& Advertising Creativity. The benchmark covers frequent real-world editing scenarios, including ID photo generation, Chinese-English text translation, virtual furniture placement, and poster generation. CPI-Intelligent-Bench consists of  1,181 image-editing reasoning instances,
covering 8 expert domains and 67 sub-disciplines as shown in Figure~\ref{fig:Intelligent_bench}. Compared with the original benchmark, ExpertVerse~\cite{wang2026expertversegeneralpurposebenchmarkexpertlevel}, CPI-Intelligent-Bench
substantially increases the dataset scale while broadening domain coverage and expanding task
granularity, thereby enabling a more comprehensive and fine-grained assessment of domain-specific
visual reasoning capabilities.





\begin{table}
    \centering
    
    \caption{Comparison of benchmarks for image editing evaluation.}
    \label{tab:benchmark_comparison}
    
    \small
    \renewcommand{\arraystretch}{1.4} 
    
    \begin{tabularx}{\linewidth}{@{} L S{1.2cm} S{1.5cm} S{2.5 cm} S{1.2cm} S{3.5 cm} @{}}
        \toprule
        \bfseries Benchmarks & \bfseries Size & \bfseries Sub-tasks & \bfseries Multi-image editing tasks & \bfseries Bilingual & \bfseries Task-Specific Evaluation Prompts \\
        \midrule
        
        ImgEdit\cite{ye2025imgedit}             & 811   & 14 & \no & \no  & \yes \\
        GEdit-Bench\cite{liu2025step1x-edit}         & 606   & 11 & \no & \yes & \no  \\
        REDEdit-Bench\cite{firered2026rededit}       & 1,673 & 15 & \no & \yes & \yes  \\
        \midrule
        
        \bfseries \mbox{CPI-General-Bench} & \bfseries 2039 & \bfseries 24 & \bfseries \yes & \bfseries \yes & \bfseries \yes  \\
        
        \bfseries \mbox{CPI-Practical-Bench}    & \bfseries 558  & \bfseries 51 & \bfseries \yes & \bfseries \yes & \bfseries \yes  \\
        \bfseries \mbox{CPI-Intelligent-Bench}    & \bfseries 1181  & \bfseries 67 & \bfseries \yes & \bfseries \yes & \bfseries \yes  \\
        
        \bottomrule
    \end{tabularx}
\end{table}
 

\begin{figure}[t]
\centering
  \includegraphics[width=1.0\textwidth]{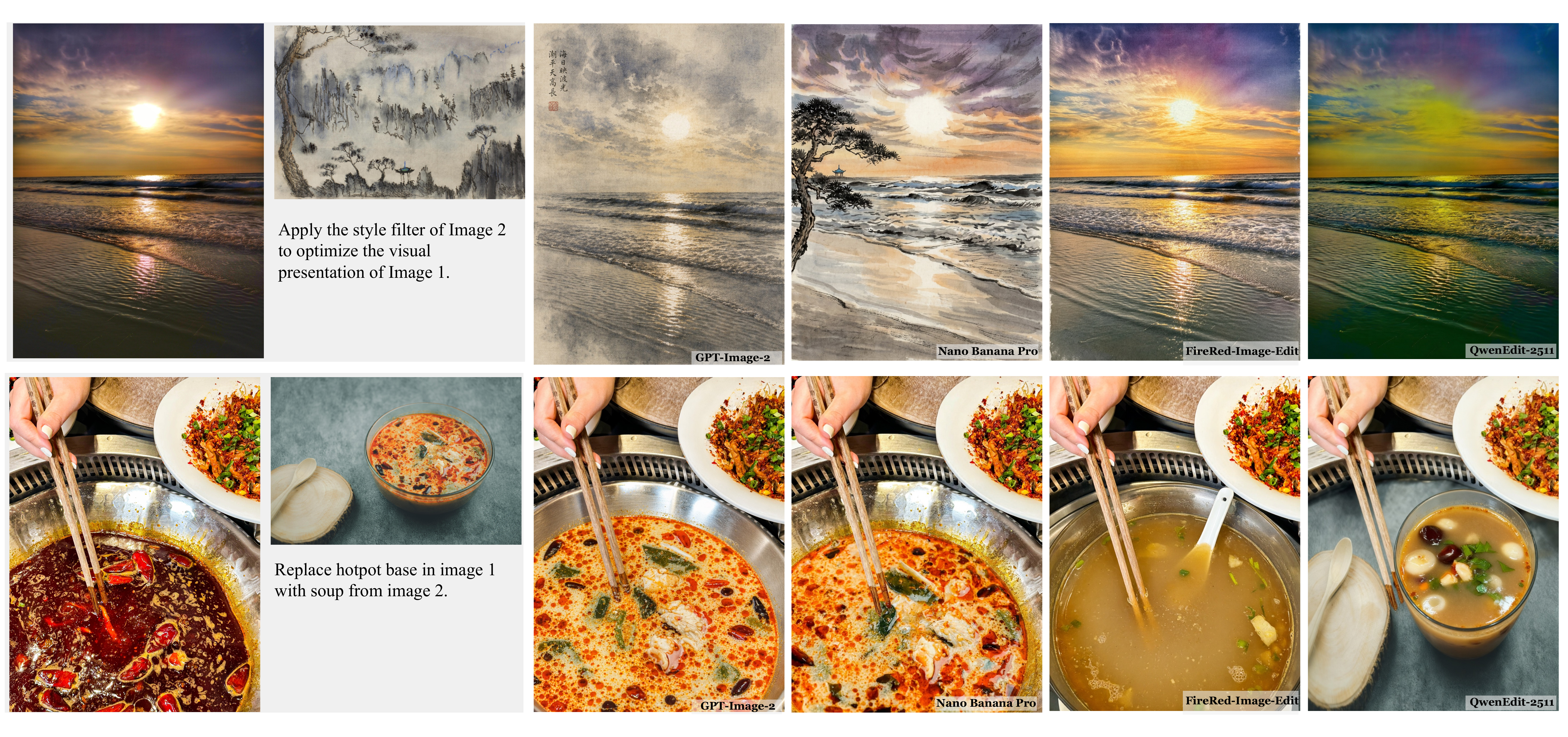}
  \caption{The results of different image editing models on CPI-General-Bench (Multi-Image Tasks)}
  \label{fig:showcase_pi_general}
\end{figure}

\begin{figure}[t]
\centering
  \includegraphics[width=1.0\textwidth]{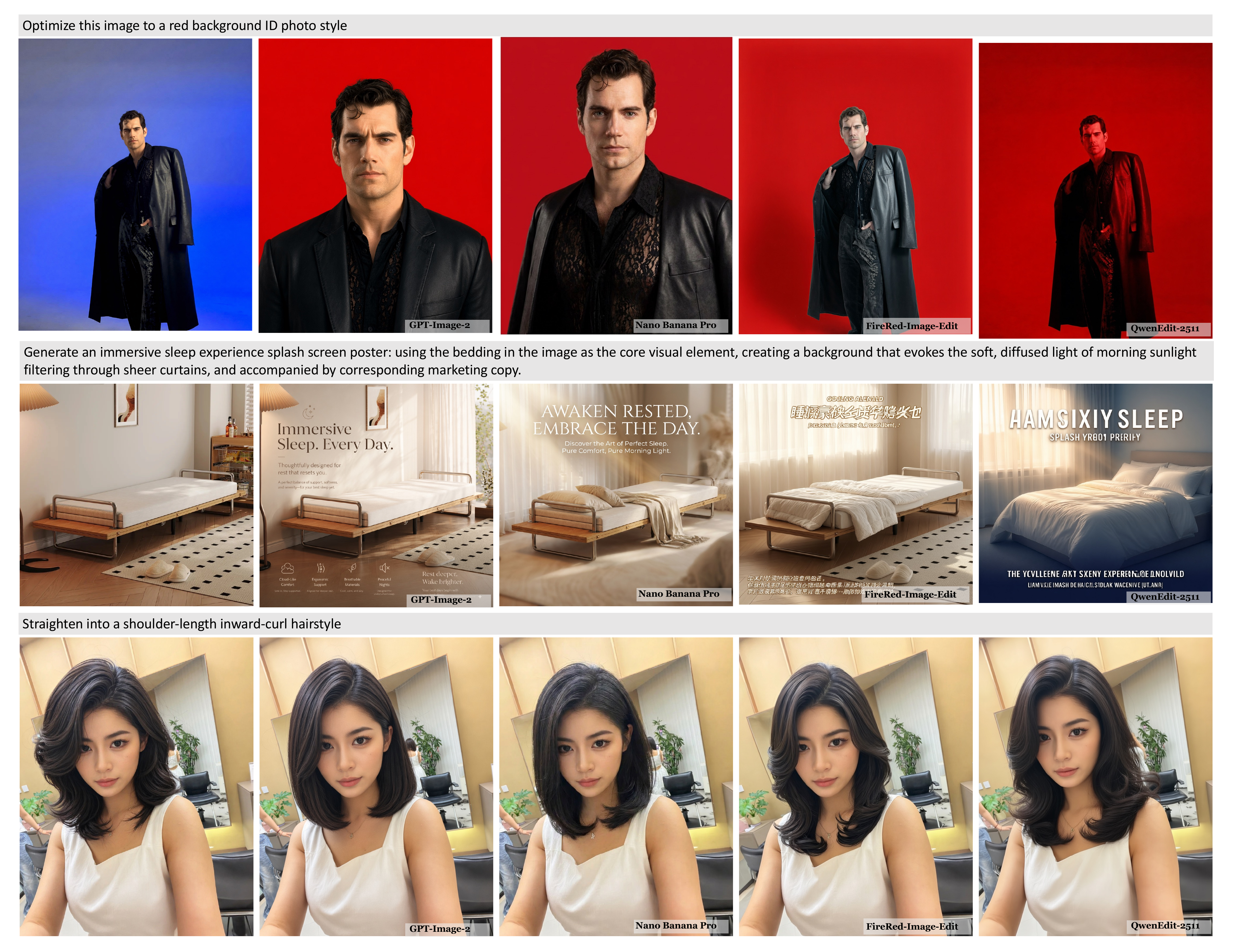}
  \caption{The results of different image editing models on CPI-Practical-Bench}
  \label{fig:showcase_pi_practical}
\end{figure}

\subsection{Evaluation Metrics}\label{section:3.3}
We established an automated evaluation framework powered by Vision-Language Models (VLMs) to evaluate CPI-General-Bench and CPI-Practical-Bench and follow the evaluation metrics in ExpertVerse~\cite{wang2026expertversegeneralpurposebenchmarkexpertlevel} for CPI-Intelligent-Bench. Specifically, for each task, we engineered specialized scoring prompts that assess edited outputs across three distinct dimensions, with scores ranging from 1 to 5. While the core metrics for most tasks focus on Instruction Adherence, Visual Naturalness, and Physical and Detail Consistency, we introduced task-specific dimensional adaptations to address unique requirements:
\textbf{(1) Text Editing Tasks:} Evaluated based on Text Instruction Compliance, Text Visual Quality, and Overall Edit Quality.
\textbf{(2) Style Transfer Tasks:} Assessed via Style Fidelity, Content Preservation, and Rendering Quality.
\textbf{(3) Identity-Consistent Tasks:} Measured using Subject Consistency, Instruction Compliance, Image Realism and Visual Coherence.
This tailored metric system ensures that the evaluation precisely captures the nuanced performance of models across different editing paradigms.

\begin{table*}[t]
    \centering
    \caption{
        Results on CPI-General-Bench-EN(Single-image editing).
    }
    \label{tab:pi_general_single}

    \begin{threeparttable}
    \small
    \renewcommand{\arraystretch}{1.2}
    \resizebox{\linewidth}{!}{
\begin{tabular}{l| c c c c c c c c c c c c c c| c}
        \toprule
        \textbf{Model}
        & \textbf{Add}
& \textbf{Adjust}
& \textbf{Compose}
& \textbf{Extract}
& \textbf{Low Level}
& \textbf{Motion}
& \textbf{Ps Human}
& \textbf{Remove}
& \textbf{Replace}
& \textbf{Stylize}
& \textbf{Structure-Guided}
& \textbf{Subject-Driven}
& \textbf{Text}
& \textbf{Viewpoint}
        & \textbf{Overall$\uparrow$}\\
        \midrule
        \rowcolor{groupgray}[0pt][0pt]
        \multicolumn{16}{@{}l@{}}{
            \hspace{0.5em}\textit{\textbf{Proprietary / Closed-source Models}}
        } \\

        GPT-Image-2\cite{gptimage2_model_card}
        & 4.56 & 4.69 & 4.69 & 4.54 & 4.72 & 4.84 & 4.64 & 4.83 & 4.77 & 4.78 & 5.00 & 4.83 & 4.53 & 4.65 & 4.74
  \\

        Nano Banana Pro\cite{google2025nanobanana}
        & 4.71 & 4.45 & 4.48 & 4.04 & 4.65 & 4.49 & 4.58 & 4.40 & 4.80 & 4.86 & 4.74 & 4.82 & 4.81 & 4.55 & 4.61
  \\

        Seedream5 Pro\cite{seedream2025seedream}
        & 4.78 & 4.65 & 4.68 & 4.10 & 4.58 & 4.79 & 4.71 & 4.56 & 4.95 & 4.84 & 4.77 & 4.84 & 4.87 & 4.45 & 4.71
  \\
        
        Seedream4.5\cite{seedream2025seedream}
        & 4.54 & 4.36 & 4.43 & 4.26 & 4.36 & 4.57 & 4.46 & 4.28 & 4.68 & 4.84 & 4.66 & 4.75 & 4.49 & 4.17 & 4.49  \\

        Qwen Image 2.0 Pro\cite{zhao2026qwen}
        & 4.70 & 4.41 & 4.47 & 3.25 & 4.44 & 4.59 & 4.57 & 4.43 & 4.80 & 4.89 & 4.37 & 4.69 & 4.71 & 4.58 & 4.55  \\
        \midrule
        \rowcolor{groupgray}[0pt][0pt]
        \multicolumn{16}{@{}l@{}}{
            \hspace{0.5em}\textit{\textbf{Open-source Models}}
        } \\

        Qwen-Image-Edit-2511~\cite{wu2025qwen}
        & 4.39 & 4.14 & 4.15 & 3.63 & 4.09 & 4.54 & 4.10 & 4.37 & 4.47 & 4.79 & 3.83 & 4.38 & 4.39 & 4.31 & 4.30  \\

        
        FLUX.2-klein-9B~\cite{blackforest2025flux2klein}
        & 4.46 & 4.14 & 3.96 & 2.46 & 3.99 & 4.08 & 4.03 & 4.31 & 4.53 & 4.83 & 4.36 & 4.62 & 3.69 & 4.62 & 4.18 \\
        
        FLUX.2-klein-4B~\cite{blackforest2025flux2klein}
        & 4.28 & 4.15 & 4.09 & 3.05 & 3.48 & 3.96 & 3.96 & 4.16 & 4.37 & 4.61 & 3.75 & 4.39 & 3.25 & 4.28 & 4.03  \\
        
        FireRed-Image-Edit~\cite{firered2026rededit}
        & 4.39 & 4.39 & 4.23 & 3.58 & 4.32 & 4.60 & 4.01 & 4.44 & 4.76 & 4.96 & 3.46 & 4.29 & 4.55 & 4.29 & 4.38  \\

        JoyAI-Image-Edit-Plus~\cite{song2026joyai}
        & 4.35 & 4.08 & 3.83 & 2.19 & 2.90 & 4.05 & 3.64 & 3.73 & 4.24 & 3.70 & 2.55 & 3.84 & 4.20 & 3.84 & 3.81  \\

        

        
        


        \bottomrule
    \end{tabular}}
    \end{threeparttable}
\end{table*}

\begin{table*}[t]
    \centering
    \caption{
        Results on CPI-General-Bench-EN(Multi-image editing).
    }
    \label{tab:pi_general_multi}

    \begin{threeparttable}
    \small
    \renewcommand{\arraystretch}{1.2}
    \resizebox{\linewidth}{!}{
\begin{tabular}{l| c c c c c c c c c c| c}
        \toprule
        \textbf{Model}
        & \textbf{Multi-Image Compose}
& \textbf{Multi-Subject-Driven}
& \textbf{Ref-Add}
& \textbf{Ref-Change}
& \textbf{Ref-Motion}
& \textbf{Ref-Remove}
& \textbf{Ref-Replace}
& \textbf{Ref-Stylize}
& \textbf{Ref-Text}
& \textbf{Ref-Viewpoint}

        & \textbf{Overall$\uparrow$}\\
        \midrule
        \rowcolor{groupgray}[0pt][0pt]
        \multicolumn{12}{@{}l@{}}{
            \hspace{0.5em}\textit{\textbf{Proprietary / Closed-source Models}}
        } \\

        GPT-Image-2\cite{gptimage2_model_card}
        & 4.65 & 4.80 & 4.56 & 4.36 & 4.20 & 4.03 & 4.60 & 4.75 & 4.42 & 4.60
        & 4.51 \\

        Nano Banana Pro\cite{google2025nanobanana}
        & 4.34 & 4.48 & 4.24 & 4.27 & 4.37 & 4.11 & 4.28 & 4.48 & 4.10 & 3.79 & 4.27  \\

        Seedream5 Pro\cite{seedream2025seedream}
        & 4.60 & 4.84 & 4.69 & 4.43 & 4.30 & 4.39 & 4.68 & 4.35 & 4.39 & 3.62 & 4.50  \\
        
        Seedream4.5\cite{seedream2025seedream}
        & 3.92 & 4.43 & 4.30 & 3.94 & 3.71 & 4.21 & 4.24 & 4.10 & 3.97 & 2.89 & 4.06  \\

        Qwen Image 2.0 Pro\cite{zhao2026qwen}
        & 4.18 & 4.56 & 4.17 & 4.19 & 3.91 & 4.03 & 4.23 & 4.25 & 3.82 & 3.46 & 4.10  \\
        \midrule
        \rowcolor{groupgray}[0pt][0pt]
        \multicolumn{12}{@{}l@{}}{
            \hspace{0.5em}\textit{\textbf{Open-source Models}}
        } \\

        Qwen-Image-Edit-2511~\cite{wu2025qwen}
        & 2.75 & 3.62 & 3.91 & 3.33 & 3.01 & 2.95 & 3.65 & 2.10 & 3.11 & 1.96 & 3.13  \\

        
        FLUX.2-klein-9B~\cite{blackforest2025flux2klein}
        & 3.68 & 3.76 & 3.93 & 3.69 & 3.01 & 3.27 & 3.76 & 3.03 & 3.11 &3.00 & 3.48  \\
        
        FLUX.2-klein-4B~\cite{blackforest2025flux2klein}
        & 3.45 & 3.65 & 4.04 & 3.70 & 2.87 & 3.03 & 3.75 & 2.88 & 2.95 & 2.97 & 3.38  \\
        
        FireRed-Image-Edit~\cite{firered2026rededit}
        & 2.58 & 3.13 & 3.83 & 3.20 & 3.47 & 3.73 & 3.21 & 2.49 & 2.76 & 2.18 & 3.11  \\

        JoyAI-Image-Edit-Plus~\cite{song2026joyai}
        & 2.70 & 3.52 & 3.78 & 3.06 & 2.51 & 2.02 & 3.41 & 1.82 & 2.93 & 2.49 & 2.89  \\

        

        
        


        \bottomrule
    \end{tabular}}
    \end{threeparttable}
\end{table*}

\begin{table*}[h]
    \centering
    \caption{
        Results on CPI-Practical-Bench-EN.
    }
    \label{tab:CPI-life-benchmark}

    \begin{threeparttable}
    \small
    \renewcommand{\arraystretch}{1.2}
    \resizebox{\linewidth}{!}{
\begin{tabular}{l| c c c c c c c c c c c c c | c}
        \toprule
         \textbf{Model}
& \textbf{Ad Create}
& \textbf{Appe. Chg}
& \textbf{Basic Edit}
& \textbf{Creative Gen.}
& \textbf{Outfit}
& \textbf{Hard Fin.}
& \textbf{Lighting}
& \textbf{Pose/View}
& \textbf{Retouch}
& \textbf{Render}
& \textbf{Soft Furn.}
& \textbf{Style and Creative Gen.}
& \textbf{Translation}
        & \textbf{Overall$\uparrow$}\\

        \midrule
        \rowcolor{groupgray}[0pt][0pt]
        \multicolumn{15}{@{}l@{}}{
            \hspace{0.5em}\textit{\textbf{Proprietary / Closed-source Models}}
        } \\

        GPT-Image-2\cite{gptimage2_model_card}
        & 4.74
        & 4.49
        & 4.42
        & 4.90
        & 4.88
        & 4.97
        & 4.80
        & 4.68
        & 4.71
        & 5.00
        & 5.00
        & 4.85
        & 4.60
        & 4.69
        \\

        Nano Banana Pro\cite{google2025nanobanana}
        & 4.90
        & 4.60
        & 4.36
        & 4.77
        & 3.94
        & 4.53
        & 5.00
        & 4.18
        & 4.61
        & 5.00
        & 4.56
        & 4.81
        & 4.27
        & 4.58
        \\

        Seedream5 Pro\cite{seedream2025seedream}
        & 4.98
        & 4.75
        & 4.67
        & 4.78
        & 4.99
        & 4.82
        & 4.80
        & 4.63
        & 4.46
        & 4.85
        & 4.79
        & 4.92
        & 4.63
        & 4.73
        \\

        Seedream4.5\cite{seedream2025seedream}
        & 4.34
        & 4.30
        & 4.09
        & 4.54
        & 4.57
        & 3.75
        & 4.80
        & 3.63
        & 4.53
        & 4.73
        & 4.40
        & 4.38
        & 3.73
        & 4.33
        \\

        Qwen Image 2.0 Pro\cite{zhao2026qwen}
        & 4.08
        & 4.48
        & 4.49
        & 4.48
        & 4.75
        & 4.38
        & 4.43
        & 4.32
        & 4.12
        & 4.80
        & 4.55
        & 4.73
        & 2.87
        & 4.39
        \\
        \midrule
        \rowcolor{groupgray}[0pt][0pt]
        \multicolumn{15}{@{}l@{}}{
            \hspace{0.5em}\textit{\textbf{Open-source Models}}
        } \\

        Qwen-Image-Edit-2511~\cite{wu2025qwen}
        & 2.73 & 3.92 & 4.17 & 4.46 & 3.38 & 3.42 & 3.70
        & 4.02 & 3.87 & 4.50 & 3.92 & 3.90 & 1.23 & 3.85 \\
        
        FLUX.2-klein-9B~\cite{blackforest2025flux2klein}
        & 2.38 & 3.94 & 4.30 & 4.27 & 3.94 & 4.05 & 2.80
        & 4.23 & 3.44 & 4.97 & 4.01 & 4.38 & 1.13 & 3.86 \\
        
        FLUX.2-klein-4B~\cite{blackforest2025flux2klein}
        & 2.10 & 3.70 & 4.05 & 3.96 & 3.70 & 3.80 & 2.10
        & 3.92 & 3.40 & 4.60 & 3.70 & 4.05 & 1.47 & 3.63 \\
        
        FireRed-Image-Edit~\cite{firered2026rededit}
        & 2.97 & 4.06 & 4.42 & 3.95 & 3.17 & 3.62 & 3.70
        & 3.73 & 3.96 & 4.88 & 3.10 & 4.18 & 1.33 & 3.89 \\

        JoyAI-Image-Edit-Plus~\cite{song2026joyai}
        & 2.76 & 3.85 & 3.68 & 2.72 & 3.73 & 3.52 & 2.33
        & 3.23 & 3.74 & 4.08 & 4.11 & 2.96 & 1.03 & 3.52 \\

        

        
        


        \bottomrule
    \end{tabular}}
    \end{threeparttable}
\end{table*}

\begin{table*}
    \centering
    \caption{
    Results on CPI-Intelligent-Bench. 
    }
    \label{tab:i2i_reasoning}

    \begin{threeparttable}
    \small
    \setlength{\tabcolsep}{6pt}
    \renewcommand{\arraystretch}{1.2}
    \begin{tabular}{l| c c c c c c c c c}
        \toprule
        \textbf{Model}
        & \textbf{IE}
        & \textbf{GH}
        & \textbf{FN}
        & \textbf{DI}
        & \textbf{FB}
        & \textbf{LC}
        & \textbf{SE}
        & \textbf{DP}
        & \textbf{Overall$\uparrow$} \\
        \midrule
        \rowcolor{groupgray}[0pt][0pt]
        \multicolumn{10}{@{}l@{}}{
            \hspace{0.5em}\textit{\textbf{Proprietary / Closed-source Models}}
        } \\

        GPT-Image-2\cite{gptimage2_model_card} & 5.00 & 4.47 & 4.86 & 4.76 & 4.82 & 4.92 & 4.38 & 4.93 & 4.77 \\
        Nano Banana Pro\cite{google2025nanobanana} & 4.84 & 4.59 & 4.58 & 4.50 & 4.81 & 4.76 & 4.37 & 4.87 & 4.68 \\
        Seedream5 Pro\cite{seedream2025seedream} & 4.83 & 4.63 & 4.75 & 4.67 & 4.68 & 4.80 & 4.26 & 4.92 & 4.70 \\
        Seedream4.5\cite{seedream2025seedream} & 4.31 & 3.47 & 4.35 & 2.92 & 4.23 & 3.98 & 3.12 & 3.96 & 3.79 \\
        Qwen Image 2.0 Pro\cite{zhao2026qwen} & 4.29 & 3.67 & 4.59 & 2.53 & 4.11 & 3.92 & 3.09 & 4.19 & 3.79 \\

        \midrule
        \rowcolor{groupgray}[0pt][0pt]
        \multicolumn{10}{@{}l@{}}{
            \hspace{0.5em}\textit{\textbf{Open-source Models}}
        } \\

        
        Qwen-Image-Edit-2511~\cite{wu2025qwen} & 3.51 & 2.46 & 2.90 & 1.54 & 3.06 & 2.48 & 1.57 & 2.88 & 2.54 \\
        FLUX.2-klein-9B~\cite{blackforest2025flux2klein} & 3.57 & 2.04 & 2.74 & 1.71 & 2.98 & 2.59 & 1.65 & 2.67 & 2.48 \\
        FLUX.2-klein-4B~\cite{blackforest2025flux2klein} & 3.45 & 1.83 & 2.52 & 1.66 & 3.01 & 2.38 & 1.58 & 2.39 & 2.33 \\
        FireRed-Image-Edit~\cite{firered2026rededit} & 3.74 & 2.31 & 2.99 & 1.66 & 3.22 & 2.68 & 1.75 & 2.91 & 2.65 \\
        JoyAI-Image-Edit-Plus~\cite{song2026joyai} & 2.79 & 1.83 & 1.96 & 1.57 & 2.46 & 2.14 & 1.54 & 2.21 & 2.07 \\

        \bottomrule

    \end{tabular}

    \begin{tablenotes}[flushleft]
        \footnotesize
        \item[*] Abbreviations: IE = IP \& Entertainment, GH = Geography \& History, FN = Film \& Narrative, DI = Data \& Information, FB = Fashion \& Beauty, LC = Life \& Consumption, SE = Science \& Education, DP = Design \& Product.
    \end{tablenotes}
    \end{threeparttable}
\end{table*}

\section{Evaluation}

We evaluated a diverse set of image editing models across existing benchmarks (GEdit-Bench \cite{liu2025step1x-edit}, ImgEdit-Bench \cite{ye2025imgedit}, REDEdit-Bench \cite{firered2026rededit}) and our proposed CPI-Bench. 


\paragraph{CPI-Bench vs. Other Benchmarks} As illustrated in Table~\ref{tab:editing_capability}, the performance variance among different models on the established benchmarks is significantly lower than that observed on CPI-Bench. This phenomenon stems from the fact that prior benchmarks focus exclusively on simple single-image editing tasks, where performance differences between state-of-the-art models have become relatively limited due to continuous iterations. In contrast, CPI-Bench incorporates complex challenges, including multi-image editing, real-world application scenarios and highly demanding reasoning-based editing. Consequently, it effectively amplifies performance discrepancies, providing a more comprehensive assessment of model capabilities and highlighting potential directions for future optimization.

A deeper analysis of model performance within the sub-components of CPI-Bench further validates our findings:

\textbf{Single-Image vs. Multi-Image Tasks (CPI-General-Bench):} As shown in Table~\ref{tab:pi_general_single}, on the single-image editing tasks within CPI-General-Bench, which predominantly consist of fundamental tasks, the performance gap between models remains marginal, with most models achieving scores around 4.0. However, a significant divergence emerges in the multi-image editing subset, as shown in Table~\ref{tab:pi_general_multi} and Figure~\ref{fig:showcase_pi_general}. In this challenging domain, only top-tier closed-source models consistently exceed a score of 4.0, whereas open-source models generally remain around the 3.0 range. This disparity reveals that an important factor contributing to the gap distinguishing open-source from closed-source models lies in their capability to handle multi-image consistency and reasoning.

\textbf{Real-World Application Performance (CPI-Practical-Bench):} Furthermore, as demonstrated in Table~\ref{tab:CPI-life-benchmark} and Figure~\ref{fig:showcase_pi_practical}, open-source models exhibit substantially lower performance on the CPI-Practical-Bench compared to their closed-source counterparts. This gap underscores the limitations of current open-source solutions in addressing the complexities of real-world business deployment scenarios, highlighting an important direction for future improvement.

\textbf{Intelligence Performance (CPI-Intelligent-Bench):} Since CPI-Intelligent-Bench evaluates editing tasks that require domain knowledge,
intent inference, and visual layout planning, it heavily relies on strong model reasoning capabilities. Such reasoning capabilities are often elicited through prompt engineer. Consequently, open-source image editing models lacking  prompt engineering  exhibit suboptimal performance on CPI-Intelligent-Bench as shown in
Table~\ref{tab:i2i_reasoning}.

\textbf{Comparison with Arena Rankings} To evaluate the fidelity of CPI-Bench to human perception, we conducted a direct comparison between the model rankings generated by various benchmarks and the rankings from the Arena Image Edit Leaderboard. As illustrated in Figure~\ref{fig:rank_compare}, the ranking trends of CPI-Bench exhibit the closest alignment with the Arena. Specifically, the rank positions are identical for all models except for FLUX.2-klein-9B, which deviates by only a single position (one rank gap). In stark contrast, other benchmarks display significantly divergent trends.
This visual observation is further corroborated by quantitative metrics. CPI-Bench achieves the highest Spearman correlation coefficients and the lowest Mean Absolute Error among all evaluated benchmarks. These results collectively indicate that CPI-Bench's assessment of model performance aligns most closely with the rankings derived from human evaluators on Arena, which demonstrates CPI-Bench achieves the highest degree of human preference alignment, validating its effectiveness as a useful proxy for public human preference trends.
\section{Conclusion}

In this work, we introduced CPI-Bench, a comprehensive framework designed to evaluate image editing models across three critical dimensions: editing capability, practical deployability, and complex reasoning proficiency. Unlike existing benchmarks confined to simple single-image tasks, CPI-Bench introduces the integration of multi-image editing, authentic real-world scenarios, and reasoning-based challenges. This design effectively bridges the gaps in evaluating complex interactions, significantly enhancing the comprehensiveness and reliability of performance assessment by breaking through the performance saturation bottleneck.
Furthermore, our rigorous validation against the Arena Image Edit Leaderboard reveals that the rankings generated by CPI-Bench exhibit the highest alignment with human preferences, demonstrating that CPI-Bench does not merely measure technical metrics but shows strong alignment with the perceptual judgments of human evaluators. Consequently, CPI-Bench serves as a valuable evaluation framework for precisely identifying model deficiencies, offering a reliable roadmap for future optimization that is grounded in public human preference trends.

\newpage
\bibliography{references}

\end{document}